\documentclass[10pt,twocolumn,letterpaper]{article}

\usepackage{cvpr}
\usepackage{times}
\usepackage{epsfig}
\usepackage{graphicx}
\usepackage{amsmath}
\usepackage{amssymb}
\usepackage{multirow}
\usepackage{caption} 
\allowdisplaybreaks

\usepackage[pagebackref=true,breaklinks=true,letterpaper=true,colorlinks,bookmarks=false]{hyperref}

\cvprfinalcopy 

\def\cvprPaperID{239} 

\ifcvprfinal\fi
\begin{document}

\title{A Perceptual Prediction Framework for Self Supervised Event Segmentation}

\author{Sathyanarayanan N. Aakur\\
University of South Florida\\
Tampa, FL, USA\\
{\tt\small saakur@mail.usf.edu}
\and
Sudeep Sarkar\\
University of South Florida\\
Tampa, FL, USA\\
{\tt\small sarkar@usf.edu}
}

\maketitle

\begin{abstract}
Temporal segmentation of long videos is an important problem, that has largely been tackled through supervised learning, often requiring large amounts of annotated training data. 
In this paper, we tackle the problem of self-supervised temporal segmentation that alleviates the need for any supervision in the form of labels (full supervision) or temporal ordering (weak supervision). 
We introduce a self-supervised, predictive learning framework that draws inspiration from cognitive psychology to segment long, visually complex videos into constituent events. Learning involves only a single pass through the training data.
We also introduce a new adaptive learning paradigm that helps reduce the effect of catastrophic forgetting in recurrent neural networks. 
Extensive experiments on three publicly available datasets - Breakfast Actions, 50 Salads, and INRIA Instructional Videos datasets show the efficacy of the proposed approach. 
We show that the proposed approach outperforms weakly-supervised and unsupervised baselines by up to $24\%$ and achieves competitive segmentation results compared to fully supervised baselines with only a single pass through the training data. 
Finally, we show that the proposed self-supervised learning paradigm learns highly discriminating features to improve action recognition. 
\end{abstract}
\vspace{-0.5em}
\section{Introduction}

Video data can be seen as a continuous, dynamic stream of visual cues encoded in terms of coherent, stable structures called ``\textit{events}''. 
Computer vision research has largely focused on the problem of recognizing and describing these events in terms of either labeled actions~\cite{kuehne2014language,karpathy2014large,kuehne2014language,aakurCRV2017,aakur2019wacv} or sentences (captioning)~\cite{aakur2019wacv,venugopalan2014translating,venugopalan2015sequence,guo2016attention,bin2016bidirectional,yao2015describing}. 
Such approaches assume that the video is already segmented into atomic, stable units sharing a semantic structure such as ``\textit{throw ball}'' or ``\textit{pour water}''. 
However, the problem of temporally localizing events in untrimmed video has not been explored to the same extent as activity recognition or captioning. 
In this work, we aim to tackle the problem of temporal segmentation of untrimmed videos into its constituent events in a self-supervised manner, without the need for training data annotations.

\begin{figure}[h]
\centering
\includegraphics[width=0.99\columnwidth]{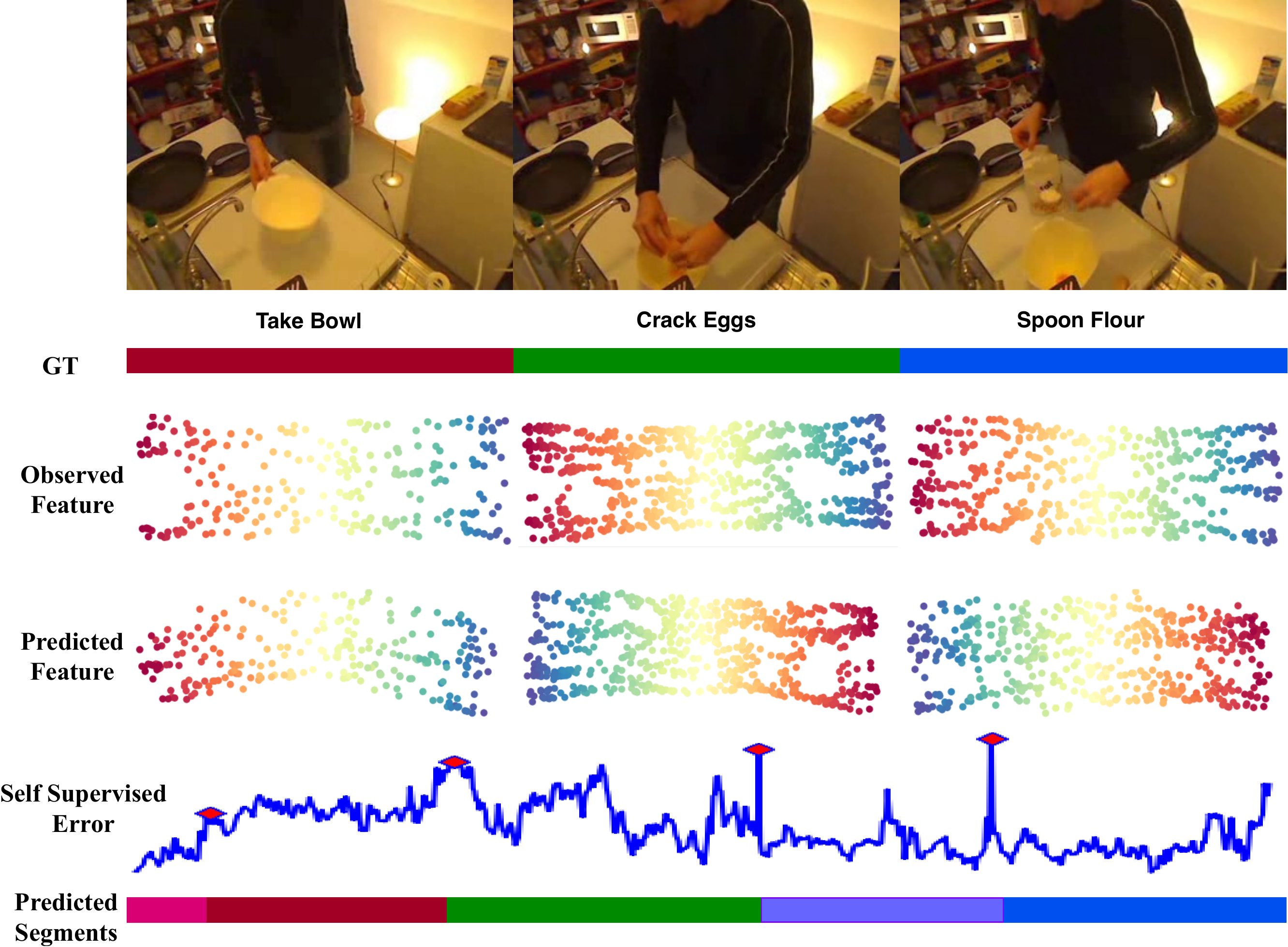}
\caption{\textbf{Proposed Approach}: Given an unsegmented input video, we encode it into a higher level feature. We predict the feature for next time instant. 
A self-supervised signal based on the difference between the predicted and the observed feature gives rise to a possible event boundary. 
} 
\label{fig:overallApproach}
\end{figure}

To segment a video into its constituent \textit{events}, we must first define the term \textit{event}. 
Drawing from cognitive psychology~\cite{zacks2001perceiving}, we define an event to be a ``\textit{segment of time at a given location that is perceived by an observer to have a beginning and an end}''. 
Event segmentation is the process of identifying these beginnings and endings and their relations. 
Based on the level of distinction, the granularity of these events can be variable. For example, \textit{throw ball} and \textit{hit ball} can be events that constitute a larger, overarching event \textit{play baseball}. 
Hence, each event can be characterized by a stable, internal representation that can be used to anticipate future visual features within the same event with high correlation, with increasing levels of error as the current transitions into the next. 
Self-supervised learning paradigms of ``\textit{predict, observe and learn}'' can then be used to provide supervision for training a computational model, typically a neural network with recurrence for temporal coherence. 

\begin{figure*}[h]
\centering
\includegraphics[width=0.99\textwidth]{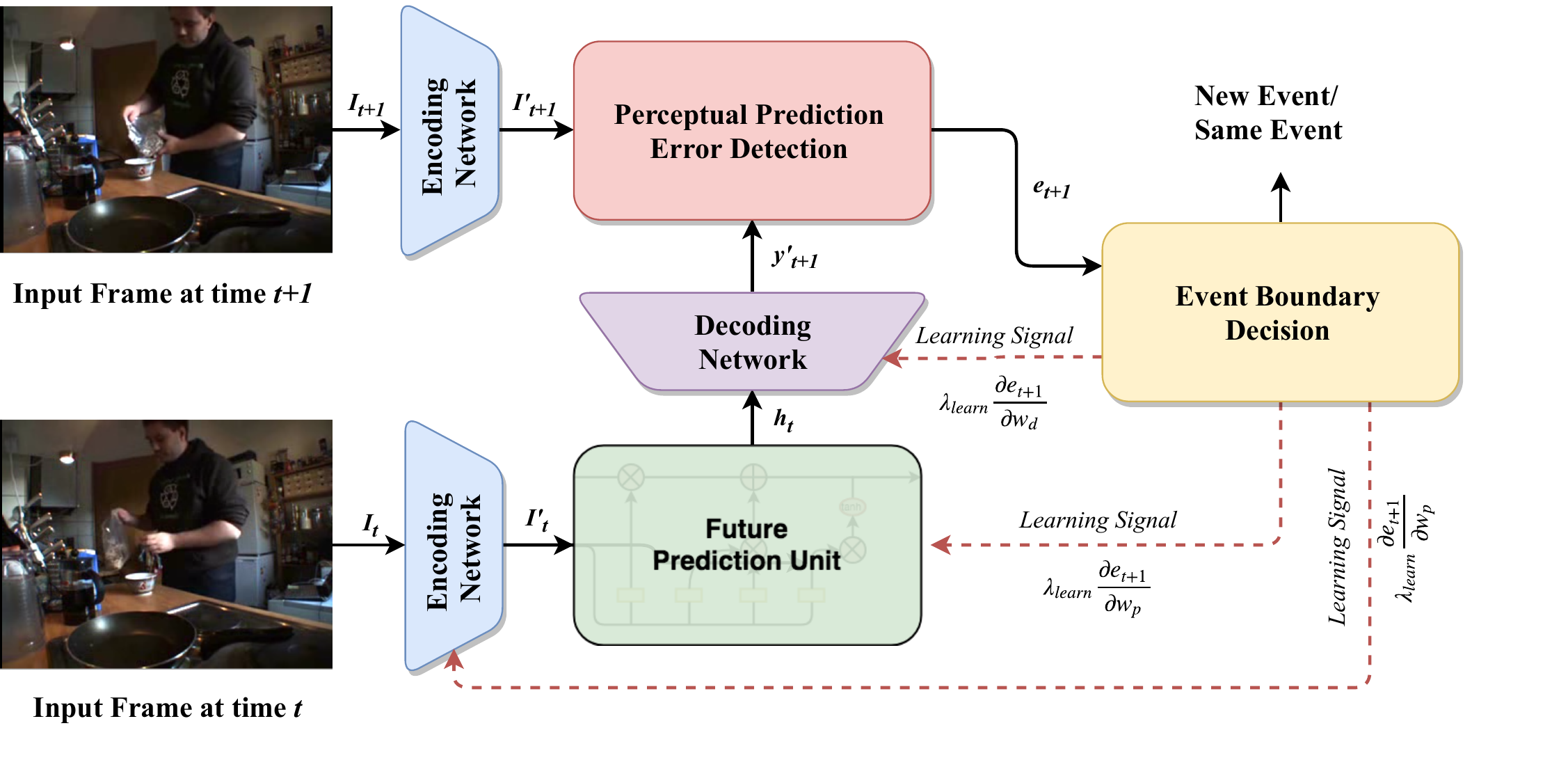}
\caption{\textbf{Overall architecture}: The proposed approach consists of four essential components: an encoder network, a predicting unit, a decoding network, and error detection and boundary decision unit.} 
\label{fig:overallArch}
\end{figure*}

We propose a novel computational model
\footnote{Additional results and code can be found at http:\textbackslash\textbackslash www.eng.usf.edu\textbackslash cvprg} 
based on the concept of perceptual prediction. Defined in cognitive psychology, it refers to the hierarchical process that transforms the current sensory inputs into state representations of the near future that allow for actions. 
Such representation of the near future enables us to anticipate sensory information based on the current event. This is illustrated in Figure \ref{fig:overallApproach}. The features were visualized using T-SNE \cite{maaten2008visualizing} for presentation.
The proposed approach has three characteristics. It is hierarchical, recurrent and cyclical. 
The hierarchical nature of the proposed approach lies in the abstraction of the incoming video frames into features of lower variability that is conducive to prediction. 
The proposed model is also recurrent. The predicted features are highly dependent on the current and previous states of the network. 
Finally, the model is highly cyclical. Predictions are compared continuously to observed features and are used to guide future predictions. 
These characteristics are common working assumptions in many different theories of perception \cite{neisser1967cognitive}, neuro-physiology \cite{fuster1991prefrontal,carpenter2016adaptive}, language processing \cite{van1983strategies} and event perception\cite{hanson1996development}.

\textbf{Contributions:} The contributions of our proposed approach are three-fold. (1) We are, to the best of our knowledge, the first to tackle the problem of self-supervised, temporal segmentation of videos. (2) We introduce the notion of self-supervised predictive learning for active event segmentation. (3) We show that understanding the spatial-temporal dynamics of events enable the model to learn the visual structure of events for better activity recognition.
\section{Related Work}
\textbf{Fully supervised approaches} treat event segmentation  as a \textit{supervised} learning problem and assign the semantics to the video in terms of labels and try to segment the video into its semantically coherent ``\textit{chunks}'', with contiguous frames sharing the same label. There have been different approaches to supervised action segmentation such as frame-based labeling using handcrafted features and a support vector machine~\cite{kuehne2014language}, modeling temporal dynamics using Hidden Markov Models~\cite{kuehne2014language}, temporal convolutional neural networks (TCN)~\cite{rene2017temporal}, spatiotemporal convolutional neural networks (CNN)~\cite{lea2016segmental} and recurrent networks~\cite{richard2017weakly} to name a few.
Such approaches often rely on the quantity and quality of the training annotations and constrained by the semantics captured in the training annotations, i.e., a closed world assumption. 

\textbf{Weakly supervised approaches} have also been explored to an extent to alleviate the need for large amounts of labeled data. 
The underlying concept behind weak supervision is to alleviate the need for direct labeling by leveraging accompanying text scripts or instructions as indirect supervision for learning highly discriminant features. 
There have been two common approaches to weakly supervised learning for temporal segmentation of videos - 
(1) using script or instructions for weak annotation\cite{bojanowski2014weakly,Ding_2018_CVPR,alayrac2016unsupervised,malmaud2015s}, and
(2) following an incomplete temporal localization of actions for learning and inference\cite{huang2016connectionist,richard2017weakly}. 
While such approaches model the temporal transitions using RNNs, they still rely on enforcing semantics for segmenting actions and hence require some supervision for learning and inference.

\textbf{Unsupervised learning} has not been explored to the same extent as supervised approaches, primarily because label semantics, if available, aid in segmentation.  
The primary approach is to use clustering as the unsupervised approach using discriminant features\cite{bhatnagar2017unsupervised,Sener_2018_CVPR}. 
The models incorporate a temporal consistency into the segmentation approach by using either LSTMs~\cite{bhatnagar2017unsupervised} or generalized mallows model ~\cite{Sener_2018_CVPR}. 
Garcia \emph{et al.}~\cite{garcia2018predicting} explore the use of a generative LSTM network to segment sequences like we do, however, they handle only coarse temporal resolution in life-log images sampled as far apart as 30 seconds. Consecutive images when events change have more variability making for easier discrimination. Besides, they require an iterative training process, which we do not.
\vspace{-0.5em}
\section{Perceptual Prediction Framework}\label{sec:general}
In this section, we introduce the proposed framework. We begin with a discussion on the perceptual processing unit, including encoding, prediction and feature reconstruction. We continue with an explanation of the self-supervised approach for training the model, followed by a discussion on boundary detection and adaptive learning. We conclude with implementation details of the proposed approach. 
It is to be noted that \cite{metcalf2017modelling} also propose a similar approach based on the Event Segmentation Theory. However, the event boundary detection is achieved using a reinforcement learning paradigm that requires significant amounts of training data and iterations and the approach has only been demonstrated on motion capture data.
\subsection{Perceptual Processing}
We follow the general principles outlined in the Event Segmentation Theory proposed by Zacks \emph{et al.}~\cite{zacks2001event,zacks2001perceiving,zacks2007event}. 
At the core of the approach, illustrated in Figure \ref{fig:overallArch} is a predictive processing platform that encodes a visual input $I(t)$ into a higher level abstraction $I^\prime(t)$ using an encoder network. 
The abstracted feature is used as a prior to predict the anticipated feature $I^\prime(t+1)$ at time $t+1$. 
The reconstruction or decoder network creates the anticipated feature, which is used to determine the event boundaries between successive activities in streaming, input video. 

\subsubsection{Visual Feature Encoding}
We encode the input frame at each time step into an abstracted, higher level visual feature and use it as a basis for perceptual processing rather than the raw input at the pixel level (for reduced network complexity) or higher level semantics (which require training data in the form of labels). 
The encoding process requires learning a function $g(I(t),\omega_e)$ that transforms an input frame $I(t)$ into a higher dimensional feature space that encodes the spatial features of the input into a feature vector $I^\prime(t)$, where $\omega_e$ is the set of learnable parameters. 
While the feature space can be pre-computed features such as Histogram of Optic Flow (HOF)~\cite{chaudhry2009histograms}, or Improved Dense Trajectories (IDT)~\cite{wang2013action}, we propose the joint training of a convolutional neural network. 

The prediction error and the subsequent error gradient described in Sections \ref{sec:errorGating} and \ref{sec:adaptiveLearning}, respectively, allow for the CNN to learn highly discriminative features, resulting in higher recognition accuracy (Section \ref{sec:improvedAcc}). 
An added advantage is that the prediction can be made at different hierarchies of feature embeddings, including at the pixel-level, allowing for event segmentation at different granularities.

\subsubsection{Recurrent Prediction for Feature Forecasting}
The prediction of the visual feature at time $t+1$ is conditioned by the observation at time $t$, $I^\prime(t)$, and an internal model of the current event. 
Formally, this can be defined by a generative model $P(I'(t+1)|\omega_p, I'(t))$, where $\omega_p$ is the set of hidden parameters characterizing the internal state of the current observed event. 
To capture the temporal dependencies among \textit{intra}-event frames and \textit{inter}-event frames, we propose the use of a recurrent network, such as recurrent neural networks (RNN) or Long Short Term Memory Networks (LSTMs)\cite{hochreiter1997long}. The predictor model can be mathematically expressed as
\begingroup
\allowdisplaybreaks
\begin{align}
\label{eqn:LSTM}
i_t = \sigma(W_{i}I^\prime(t) + W_{hi}h_{t-1} + b_i) \\ \notag
f_t = \sigma(W_{f}I^\prime(t) + W_{hf}h_{t-1} + b_f ) \\ \notag
o_t = \sigma(W_{o}I^\prime(t) + W_{ho}h_{t-1} + b_o ) \\ \notag
g_t = \phi(W_{g}I^\prime(t) + W_{hg}h_{t-1} + b_g ) \\ \notag
m_t = f_t \cdot m_{t-1} + i_t \cdot g_t \\ \notag
h_t = o_t \cdot \phi(m_t) \notag
\end{align}
\endgroup
where $\sigma$ is a non-linear activation function, the dot-operator ($\cdot$) represents element-wise multiplication, $\phi$ is the hyperbolic tangent function (\emph{tanh}) and $W_x$ and $b_x$ represent the trained weights and biases for each of the gates. Collectively, $\{W_{hi}, W_{hf}, W_{ho}, W_{hg}\}$ and their respective biases constitute the learnable parameters $\omega_p$.

As can be seen from Equation~\ref{eqn:LSTM}, there are four common ``gates'' or layers that help the prediction of the network - the input gate $i_t$, forget gate $f_t$, output gate $o_t$, the memory layer $g_t$, the memory state $m_t$ and the event state $h_t$. 
In the proposed framework, the memory state $m_t$ and the event state $h_t$ are key to the predictions made by the recurrent unit. 
The event state $h_t$ is a representation of the event observed at time instant $t$ and hence is sensitive to the observed input $I^\prime(t)$ than the event layer, which is more persistent across events. 
The event layer is a gated layer, which receives input from the encoder as well as the recurrent event model. 
However, the inputs to the event layer are modulated by a self-supervised gating signal (Section~\ref{sec:errorGating}), which is indicative of the quality of predictions made by the recurrent model. 
The gating allows for updating the weights quickly but also maintains a coherent state within the event. 

\textbf{Why recurrent networks?} While convolutional decoder networks~\cite{jia2016dynamic} and mixture-of-network models~\cite{vondrick2016anticipating} are viable alternatives for future prediction, we propose the use of recurrent networks for the following reasons. 
Imagine a sequence of frames $I_a = (I_a^1, I_a^2, \ldots I_a^n)$ corresponding to the activity $a$. Given the complex nature of videos such as those in instructional or sports domains, the next set of frames can be followed by frames of activity $b$ or $c$ with equal probability, given by $I_b = (I_b^1, I_b^2, \ldots I_b^m)$ and $I_c = (I_c^1, I_c^2, \ldots I_c^k)$ respectively. Using a fully connected or convolutional prediction unit is likely to result in the prediction of features that tend to be the average of the two activities $a$ and $b$, i.e. $I_{avg}^k = \frac{1}{2}(I_b^k + I_c^k)$ for the time $k$. 
This is not a desirable outcome because the predicted features can either be an unlikely outcome or, more probably, be outside the plausible manifold of representations. 
The use of recurrent networks such as RNNs and LSTMs allow for multiple futures that can be possible at time $t+1$, conditioned upon the observation of frames until time $t$.

\subsubsection{Feature Reconstruction}
In the proposed framework, the goal of the perceptual processing unit (or rather the reconstruction network) is to reconstruct the predicted feature $y^\prime_{t+1}$ given a source prediction $h_t$, which maximizes the probability

\begin{equation}
p(y^\prime_{t+1} | h_t) \propto p(h_t|y^\prime_{t+1})\,p(y^\prime_{t+1})
\label{eqn:predictedFeat}
\end{equation}
where the first term is the likelihood model (or translation model from NLP) and the second is the feature prior model. 
However, we model $\log\,p(y^\prime_{t+1} | h_t)$ as a log-linear model $f(\cdot)$ conditioned upon the weights of the recurrent model $\omega_p$ and the observed feature $I^\prime(t)$ and characterized by 
\begin{equation}
\log\,p(y^\prime_{t+1} | h_t) = 
\sum_{n=1}^{t} f(\omega_p, I^\prime(t)) + log\,Z(h_t)
\label{eqn:predictedFeat1}
\end{equation}
where $Z(h_t)$ is a normalization constant that does not depend on the weights $\omega_p$. The reconstruction model completes the generative process for forecasting the feature at time $t+1$ and helps in constructing the self-supervised learning setting for identifying event boundaries.

\subsection{Self-Supervised Learning}\label{sec:Error}
The quality of the predictions is determined by comparing the prediction from the predictor model $y^\prime(t)$ to the observed visual feature $I^\prime(t)$. 
The deviation of the predicted input from the observed features is termed as the perceptual prediction error $E_P(t)$ and is described by the equation:
\begin{equation}
E_P(t) = \sum_{i=1}^{n} {\lVert I^\prime(t) - y^{\prime}(t)\lVert_{\ell_1}^2}
\label{EPLoss}
\end{equation}
where $E_P(t)$ is the perceptual prediction error at time $t$, given the predicted visual $y^{\prime}(t)$ and the actual observed feature at time $t$, $I^\prime(t)$. The predicted input is obtained through the inference function defined in Equation \ref{eqn:predictedFeat}. 
The perceptual prediction error is indicative of the prediction quality and is directly correlated with the quality of the recurrent model's internal state $h(t)$. 
Increasingly large deviations indicate that the current state is not a reliable representation of the observed event. 
Hence, the gating signal serves as an indicator of event boundaries. 
The minimization of the perceptual prediction error serves as the objective function for the network during training.

\subsection{Error Gating for Event Segmentation}\label{sec:errorGating}
The gating signal (Section \ref{sec:general}) is an integral component in the proposed framework. 
We hypothesize that the visual features of successive events differ significantly at event boundaries. 
The difference in visual features can be minor among sub-activities and can be large across radically different events. 
For example, in Figure~\ref{fig:overallApproach}, we can see that the visual representation of the features learned by the encoder network for the activities \textit{take bowl} and \textit{crack eggs} are closer together than the features between the activities \textit{take bowl} and \textit{spoon flour}. 
This diverging feature space causes a transient increase in the perceptual prediction error, especially at event boundaries. 
The prediction error decreases as the predictor model adapts to the new event.
%
This is illustrated in Figure \ref{fig:overallApproach}. We show the perceptual prediction error (second from the bottom) and the ground truth segmentation (second from the top) for the video “Make Pancake”. 
As illustrated, the error rates are higher at the event boundaries and lower among ``in-event'' frames.

The unsupervised gating signal is achieved using an anomaly detection module. In our implementation, we use a low pass filter. The low pass filter maintains a relative measure of the perceptual prediction error made by the predictor module. It is a relative measure because the low pass filter only maintains a running average of the prediction errors made over the last n time steps. The perceptual quality metric, $P_q$, is given by:

\begin{equation}
P_q(t) = P_q(t-1) + \frac{1}{n} (E_P(t) - P_q(t-1))
\end{equation}
where $n$ is the prediction error history that influences the anomaly detection module's internal model for detecting event boundaries. In our experiments, we maintain $n$ at 5. 
This is chosen based on the average response time of human perception, which is around 200 ms \cite{thorpe1996speed}. 

The gating signal, $G(t)$, is triggered when the current prediction error exceeds the average quality metric by at least $50\%$.

\begin{equation}
 G(t) = 
\begin{cases}
    1, & \frac{E_P(t)}{P_q(t-1)} > \psi_{e} \\
    0, & \text{otherwise}
\end{cases}
\end{equation}
where $P_E(t)$ is the perceptual prediction error at time $t$, $G(t)$ is the value of the gating signal at time $t$, $P_q(t-1)$ is the prediction quality metric at time $t$ and $\Psi_e$ is the prediction error threshold for boundary detection. 
For optimal prediction, the perceptual prediction error would be very high at the event boundary frames and very low at all within-event frames. In our experiments, $\Psi_e$ is set to be $1.5$.

In actual, real-world video frames, however, there exist additional noise in the form of occlusions and background motion which can cause some event boundaries to have a low perceptual prediction error. 
In that case, however, the gating signal would continue to be low and become high when there is a transient increase in error. 
This is visualized in Figure~\ref{fig:overallApproach}. It can be seen that the perceptual errors were lower at event boundaries between activities \textit{take bowl} and \textit{crack eggs} in a video of ground truth \textit{make pancakes}.
However, the prediction error increases radically soon after the boundary frames, indicating a new event. Such cases could, arguably, be attributed to conditions when there are lesser variations in the visual features at an event boundary.

\subsection{Adaptive Learning for Plasticity}\label{sec:adaptiveLearning}
The proposed training of the prediction module is particularly conducive towards overfitting since we propagate the perceptual prediction error at every time step. This introduces severe overfitting, especially in the prediction model. 
To allow for some plasticity and avoid catastrophic forgetting in the network, we introduce the concept of adaptive learning. 
This is similar to the learning rate schedule, a commonly used technique for training deep neural networks. 
However, instead of using predetermined intervals for changing the learning rates, we propose the use of the gating signal to modulate the learning rate. 
For example, when the perceptual prediction rate is lower than the average prediction rate, the predictor model is considered to have a good, stable representation of the current event. 
Propagating the prediction error, when there is a good representation of the event can lead to overfitting of the predictor model to that particular event and does not help generalize. 
Hence, we propose lower learning rates for time steps when there are negligible prediction error and a relatively higher (by a magnitude of 100) for when there is higher prediction error. 
Intuitively, this adaptive learning rate allows the model to adapt much quicker to new events (at event boundaries where there are likely to be higher errors) and learn to maintain the internal representation for within-event frames. 

Formally, the learning rate is defined as the result of the adaptive learning rule defined as a function of the perceptual prediction error defined in Section \ref{sec:Error} and is defined as 
\begin{equation}
 \lambda_{learn} = 
\begin{cases}
    \Delta^{-}_{t}\lambda_{init}, & E_P(t) > \mu_{e}\\
    \Delta^{+}_{t}\lambda_{init}, & E_P(t) < \mu_{e}\\
    \lambda_{init}, & otherwise
\end{cases}
\label{eqn:learnRate}
\end{equation}
where $\Delta^{-}_{t}$, $\Delta^{+}_{t}$ and $\lambda_{init}$ refer to the scaling of the learning rate in the negative direction, positive direction and the initial learning rate respectively and $\mu_{e} = \frac{1}{t_2 -t_1} \int_{t_1}^{t_2} E_P\, dE_P$. 
The learning rate is adjusted based on the quality of the predictions characterized by the perceptual prediction error between a temporal sequence between times $t_1$ and $t_2$, typically defined by the gating signal..
The impact of the adaptive changes to the learning rate is shown in the quantitative evaluation Section \ref{ablative}, where the adaptive learning scheme shows improvement of up to 20\% compared to training without the learning scheme.

\begin{figure*}[h]
\centering
\includegraphics[width=0.95\textwidth]{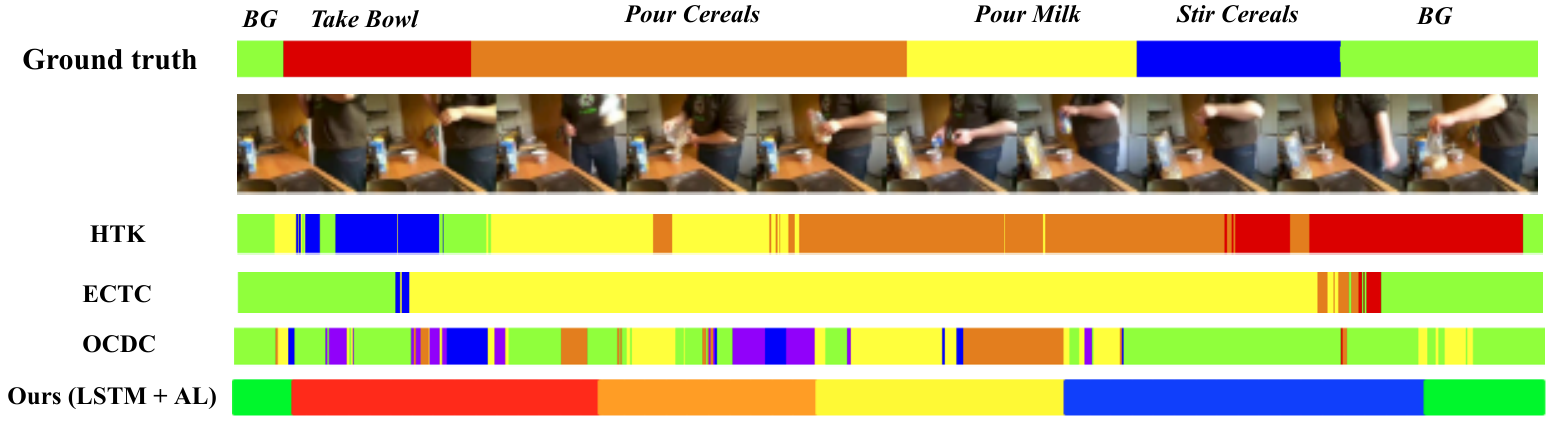}
\caption{Illustration of the segmentation performance of the proposed approach on the Breakfast Actions Dataset on a video with ground truth \textit{Make Cereals}. The proposed approach does not show the tendency to over-segment and provides coherent segmentation. The approach, however, shows a tendency to take longer to detect boundaries for visually similar activities.} 
\label{fig:segmentationBA}
\end{figure*}

\subsection{Implementation Details}
In our experiments, we use a VGG-16~\cite{simonyan2014very} network pre-trained on ImageNet as our hierarchical, feature encoder module. 
We discard the final layer and use the second fully connected layer with 4096 units as our encoded feature vector for a given frame. 
The feature vector is then consumed by a predictor model. We trained two versions, one with an RNN and the other with an LSTM as our predictor models. 
The LSTM model used is the original version proposed by~\cite{hochreiter1997long}. 
Finally, the anomaly detection module runs an average low pass filter described in Section~\ref{sec:errorGating}. 
The initial learning rate described in Section~\ref{sec:adaptiveLearning} is set to be $1\times10^{-6}$. The scaling factors $\Delta^{-}_{t}$ and $\Delta^{+}_{t}$ are set to be $1\times10^{-2}$ and $1\times10^{-3}$, respectively.
The training was done on a computer with one Titan X Pascal. 
\section{Experimental Evaluation}
\subsection{Datasets}
We evaluate and analyze the performance of the proposed approach on three large, publicly available datasets - Breakfast Actions \cite{kuehne2014language}, INRIA Instructional Videos dataset\cite{alayrac2016unsupervised} and the 50 Salads dataset \cite{stein2013combining}. 
Each dataset offers a different challenge to the approach allow us to evaluate its performance on a variety of challenging conditions. 

\textbf{Breakfast Actions Dataset} is a large collection of 1,712 videos of 10 breakfast activities performed by 52 actors. Each activity consists of multiple sub-activities that possess visual and temporal variations according to the subject's preferences and style. Varying qualities of visual data as well as complexities such as occlusions and viewpoints increase the complexity of the temporal segmentation task.

\textbf{INRIA Instructional Videos Dataset} contains 150 videos of 5 different activities collected from YouTube. Each of the videos are, on average, 2 minutes long and have around 47 sub-activities. There also exists a ``background activities'' which consists of sequence where there does not exist a clear sub-activity that is visually discriminable. This offers a considerable challenge for approaches that are not explicitly trained for such visual features.

\textbf{50 Salads Dataset} is a multimodal data collected in the cooking domain. The datasets contains over four (4) hours of annotated data of 25 people preparing 2 mixed salads each and provides data in different modalities such as RGB frames, depth maps and accelerometer data for devices attached to different items such as knives, spoons and bottles to name a few. The annotations of activities are provided at different levels of granularities - high, low and eval. We use the ``eval'' granularity following evaluation protocols in prior works~\cite{lea2016segmental,rene2017temporal}.

\subsection{Evaluation Metrics}
We use two commonly used evaluation metrics for analyzing the performance of the proposed model. 
We used the same evaluation protocol and code as in~\cite{alayrac2016unsupervised, Sener_2018_CVPR}. 
We used the Hungarian matching algorithm to obtain the one-to-one mappings between the predicted segments and the ground truth to evaluate the performance due to the unsupervised nature of the proposed approach. 
We use the mean over frames (MoF) to evaluate the ability of the proposed approach to temporally localize the sub-activities. We evaluate the divergence of the predicted segments from the ground truth segmentation using the Jaccard index (Intersection over Union or IoU). 
We also use the F1 score to evaluate the quality of the temporal segmentation. The evaluation protocol for the recognition task in Section \ref{sec:improvedAcc} is the unit level accuracy for the 48 classes as seen in Table 3 from \cite{kuehne2014language} and compared in \cite{kuehne2014language, aakur2019wacv,desouza2016spatially,huang2016connectionist}.

\subsection{Ablative Studies} We evaluate different variations of our proposed approach to compare the effectiveness of each proposed component. 
We varied the prediction history $n$ and the prediction error threshold $\Psi$. 
Increasing frame window tends to merge frames and smaller clusters near the event boundaries to the prior activity class due to transient increase in error. 
This results in higher IoU and lower MoF. Low error threshold results in over segmentation as boundary detection becomes sensitive to small changes. The number of predicted clusters decreases as the window size and threshold increases.
We also trained four (4) models, with different predictor units. We trained two recurrent neural networks (RNN) as the predictor units with and without adaptive learning described in Section~\ref{sec:adaptiveLearning} indicated as \textit{RNN + No AL} and \textit{RNN + AL}, respectively. We also trained LSTM without adaptive learning (\textit{LSTM + No AL}) to compare against our main model (\textit{LSTM + AL}). We use RNNs as a possible alternative due to the short-term future predictions (1 frame ahead) required.
We discuss these results next.

\subsection{Quantitative Evaluation}
\begin{figure*}[h]
\centering
\includegraphics[width=0.95\textwidth]{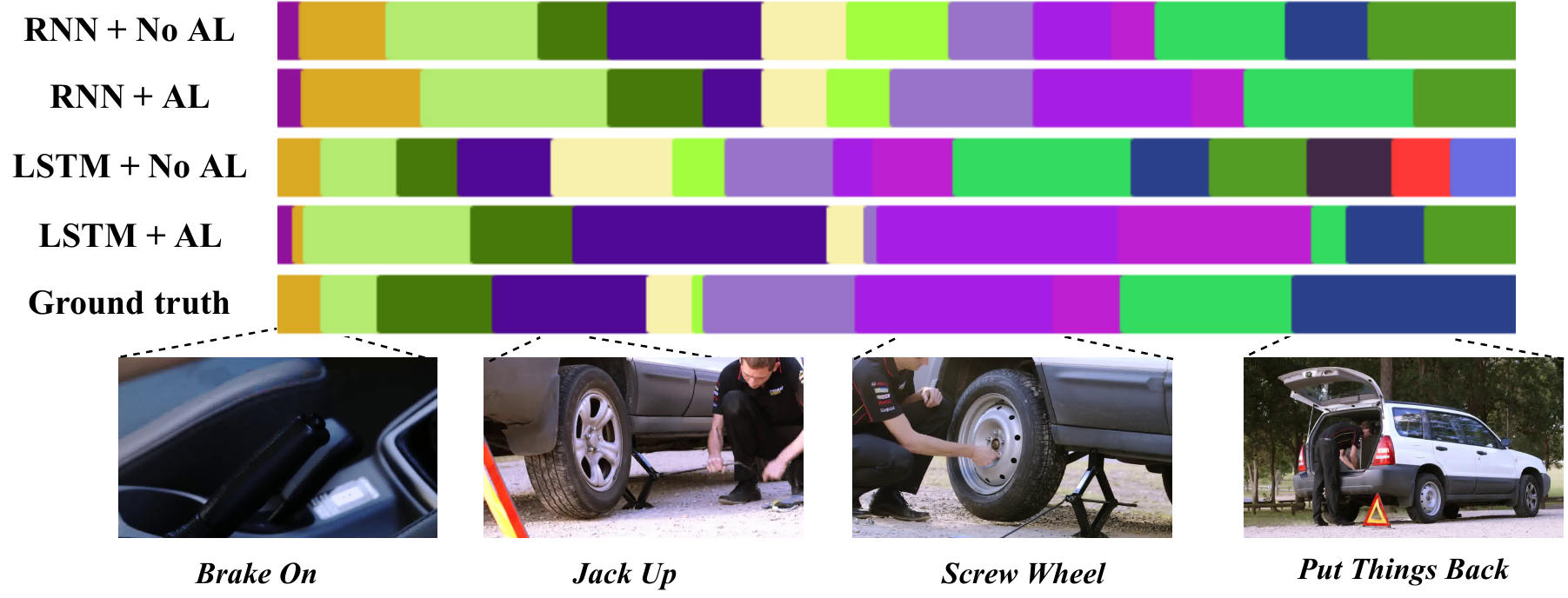}
\caption{\textbf{Ablative Studies}: Illustrative comparison of variations of our approach, using RNNs and LSTMs with and without adaptive learning on the INRIA Instructional Videos Dataset on a video with ground truth \textit{Change Tire}. It can be seen that complex visual scenes with activities of shorter duration pose a significant challenge to the proposed framework and cause fragmentation and over segmentation. However, the use of adaptive learning helps alleviate this to some extent. Note: Temporal segmentation time lines are shown without the background class for better visualization.} 
\label{fig:segmentationINRIA}
\end{figure*}

\textbf{Breakfast Actions Dataset}
We evaluate the performance of our full model \textit{LSTM + AL} on the breakfast actions dataset and compare against fully supervised, weakly supervised and unsupervised approaches. We show the performance of the SVM\cite{kuehne2014language} approach to highlight the importance of temporal modeling. As can be seen from Table \ref{ba_Performance_Table}, the proposed approach outperformed all unsupervised and weakly supervised approaches, and some fully supervised approaches.

\begin{table}[h]
\centering
\begin{tabular}{|c|c|c|c|}
\hline
\textbf{Supervision} & \textbf{Approach}     & \textbf{MoF} & \textbf{IoU} \\ \hline
\multirow{5}{*}{Full} & SVM \cite{kuehne2014language} & 15.8 & - \\
& HTK(64)\cite{kuehne2016end} & 56.3 & - \\
& ED-TCN\cite{rene2017temporal} & 43.3 & 42.0 \\
& TCFPN\cite{Ding_2018_CVPR} & 52.0 & \textbf{54.9} \\
& GRU\cite{richard2017weakly} & \textbf{60.6} & - \\
\hline
\multirow{4}{*}{Weak} & OCDC\cite{bojanowski2014weakly} & 8.9 & 23.4 \\
& ECTC\cite{huang2016connectionist} & 27.7 & - \\
& Fine2Coarse\cite{richard2016temporal} & 33.3 & \textbf{47.3} \\
& TCFPN + ISBA\cite{Ding_2018_CVPR} & \textbf{38.4} & 40.6 \\\hline
\multirow{2}{*}{None} & KNN+GMM\cite{Sener_2018_CVPR} & 34.6 & \textbf{47.1} \\
& Ours (LSTM + AL) & \textbf{42.9} & 46.9 \\\hline
\end{tabular}
\caption{Segmentation Results on the Breakfast Action dataset. MoF refers to the Mean over Frames metric and IoU is the Intersection over Union metric.}
\label{ba_Performance_Table}
\end{table}

It should be noted that the other unsupervised approach~\cite{Sener_2018_CVPR}, requires the number of clusters (from ground truth) to achieve the performance whereas our approach does not require such knowledge and is done in a streaming fashion.
Additionally, the weakly supervised methods~\cite{huang2016connectionist,richard2016temporal,Ding_2018_CVPR} require both the number of actions as well as an ordered list of
sub-activities as input. 
ECTC~\cite{huang2016connectionist} is based on discriminative
clustering, while OCDC~\cite{bojanowski2014weakly} and Fine2Coarse~\cite{richard2016temporal} are also RNN-based methods.

\textbf{50 Salads Dataset}
We also evaluate our approach on the 50 Salads dataset, using only the visual features as input. We report the Mean of Frames (MoF) metric for fair comparison. 
As can be seen from Table \ref{50s_Performance_Table}, the proposed approach significantly outperforms the other unsupervised approach, improving by over $11\%$. We also show the performance of the frame-based classification approaches VGG and IDT \cite{lea2016segmental} to show the impact of temporal modeling.
\begin{table}[h]
\centering
\begin{tabular}{|c|c|c|}
\hline
\textbf{Supervision} & \textbf{Approach}     & \textbf{MoF} \\ \hline
\multirow{6}{*}{Full} 
& VGG**\cite{lea2016segmental} & 7.6\% \\
& IDT**\cite{lea2016segmental} & 54.3\% \\
& S-CNN + LSTM\cite{lea2016segmental} & 66.6\% \\
& TDRN\cite{lei2018temporal} & 68.1\% \\
 & ST-CNN + Seg\cite{lea2016segmental}  & 72.0\% \\
 & TCN\cite{rene2017temporal} & \textbf{73.4}\% \\
\hline
\multirow{2}{*}{None}
& LSTM + KNN\cite{bhatnagar2017unsupervised} & 54.0\% \\
& Ours (LSTM + AL) & \textbf{60.6\%}   \\
\hline
\end{tabular}
\caption{Segmentation Results on the 50 Salads dataset, at granularity `\textit{Eval}`. **Models were intentionally reported without temporal constraints for ablative studies. }
\label{50s_Performance_Table}
\end{table}
It should be noted that the fully supervised approaches required significantly more training data - both in the form of labels as well as training epochs. 
Additionally, the TCN approach~\cite{rene2017temporal} uses the accelerometer data as well to achieve the state-of-the-art performance of $74.4\%$

\textbf{INRIA Instructional Videos Dataset:}\label{ablative}
Finally, we evaluate our approach on the INRIA Instructional Videos dataset, which posed a significant challenge in the form of high amounts of background (noise) data. We report the F1 score for fair comparison to the other state-of-the-art approaches.
As can be seen from Table \ref{INRIA_Performance_Table}, the proposed model outperforms the other unsupervised approach \cite{Sener_2018_CVPR} by $23.3\%$, the weakly supervised approach \cite{bojanowski2014weakly} by $24.8\%$ and has competitive performance to the fully supervised approaches\cite{malmaud2015s,alayrac2016unsupervised,Sener_2018_CVPR}.

\begin{table}[h]
\centering
\begin{tabular}{|c|c|c|c|}
\hline
\textbf{Supervision} & \textbf{Approach} & \textbf{F1} \\ \hline
\multirow{3}{*}{Full} 
& HMM + Text \cite{malmaud2015s}  & 22.9\% \\
&  Discriminative Clustering\cite{alayrac2016unsupervised} & 41.4\% \\
& KNN+GMM\cite{Sener_2018_CVPR} + GT  & \textbf{69.2\%} \\
\hline
\multirow{2}{*}{Weak} 
& OCDC + Text Features \cite{bojanowski2014weakly}  & 28.9\% \\ 
& OCDC \cite{bojanowski2014weakly}  & \textbf{31.8\%} \\ 
\hline
\multirow{5}{*}{None} & KNN+GMM\cite{Sener_2018_CVPR}  & 32.2\%  \\
& Ours (RNN + No AL)  & 25.9\% \\
& Ours (RNN + AL)  & 29.4\% \\
& Ours (LSTM + No AL)  & 36.4\% \\
& Ours (LSTM + AL)  & \textbf{39.7}\% \\
\hline
\end{tabular}
\caption{Segmentation Results on the INRIA Instructional Videos dataset. We report F1 score for fair comparison.}
\label{INRIA_Performance_Table}
\end{table}
We also evaluate the performance of the models with and without adaptive learning. It can be seen that long term temporal dependence captured by LSTMs is significant, especially due to the long durations of activities in the dataset. Additionally, the use of adaptive learning has a significant improvement in the segmentation framework, improving the performance by $9\%$ and $11\%$ for the RNN-based model and the LSTM-based model respectively, indicating a reduced overfitting of the model to the visual data.

\subsubsection{Improved Features for Action Recognition}\label{sec:improvedAcc}
To evaluate the ability of the network to learn highly discriminative features for recognition, we evaluated the performance of the proposed approach in a recognition task. 
We use the model pretrained on the segmentation task on the Breakfast Actions dataset and use the hidden layer of the LSTM as input to a fully connected layer and use cross entropy to train the model. 
We also trained another network with the same structure - VGG16 + LSTM without the pretraining on the segmentation task to compare the effect of the features learned using self-supervision.
\begin{table}[ht]
\centering
\begin{tabular}{|c|c|}
\hline
\textbf{Approach}     & \textbf{Precision} \\ \hline
HCF + HMM \cite{kuehne2014language}                   & 14.90\%              \\ 
HCF + CFG + HMM \cite{kuehne2014language}             & 31.8\%               \\ 
RNN + ECTC \cite{huang2016connectionist} & 35.6\% \\
RNN + ECTC (Cosine) \cite{huang2016connectionist} & 36.7\% \\
HCF + Pattern Theory \cite{desouza2016spatially}  & 38.6\%              \\ 
HCF + Pattern Theory + ConceptNet\cite{aakur2019wacv} & 42.9\%              \\ 
VGG16 + LSTM & 33.54\%              \\
VGG16 + LSTM + Predictive Features(AL) & 37.87\%              \\
\hline
\end{tabular}
\caption{Activity recognition results on Breakfast Actions dataset. HCF and AL refer to handcrafted features and Adaptive Learning, respectively.}
\label{BA_Actions_Performance_Table}
\end{table}
As can be seen from Table \ref{BA_Actions_Performance_Table}, the use of self-supervision to pretrain the network prior to the recognition task improves the recognition performance of the network and has comparable performance to the other state-of-the-art approaches. It improves the recognition accuracy by $13.12\%$ over the network without predictive pretraining.
\subsection{Qualitative Evaluation}
Through the predictive, self supervised framework, we are able to learn the sequence of visual features in streaming video. We visualize the segmentation performance of the proposed framework on the Breakfast Actions Dataset in Figure \ref{fig:segmentationBA}. 
It can be seen that the proposed approach has high temporal coherence and does not suffer from over segmentation, especially when the segments are long. Long activity sequences allow the model to learn from observation by providing more samples of ``\textit{intra-}event'' samples. 
Additionally, it can be seen that weakly supervised approaches like OCDC\cite{bojanowski2014weakly} and ECTC\cite{huang2016connectionist} suffer from over segmentation and intra-class fragmentation. 
This could arguably be attributed to the fact that they tend to enforce semantics, in the form of weak ordering of activities in the video regardless of the changes in visual features. 
Fully supervised approaches, such as HTK\cite{kuehne2016end} perform better, especially due to the ability to assign semantics to visual features. 
However, they are also affected by unbalanced data and dataset shift, as can be seen in Figure \ref{fig:segmentationBA} where the background class was segmented into other classes.

We also qualitatively evaluated the impact of adaptive learning and long term temporal memory in Figure \ref{fig:segmentationINRIA}, where the performance of the alternative methods described in Section \ref{ablative}. 
It can be seen that the use of adaptive learning during training allows the model to not overfit to any single class' intra-event frames and help generalize to other classes regardless of amount of training data. 
It is not to say that the problem of unbalanced data is alleviated, but the adaptive learning \textit{does} help to some extent. 
It is interesting to note that the LSTM model tends to over-segment when not trained with adaptive learning, while the RNN-based model does not suffer from the same fate.

\section{Conclusion}
We demonstrate how a self-supervised learning paradigm can be used to segment long, highly complex visual sequences.  There are key differences between our approach and fully supervised or weakly supervised approaches, including classical ones such as DBMs and HMMs. At a high level, our approach is unsupervised and does not require labeled data for training. The predictive error serves as supervision for training the framework. The other major aspect is that our approach requires only a single pass through the training data. Hence, the training time is very low. The experimental results demonstrate the robustness, high performance, and the generality of the approach on multiple real world datasets.


{\small
\bibliographystyle{ieee}
\bibliography{egbib}
}

\end{document}